\documentclass[11pt,onecolumn,letterpaper]{article}

\usepackage{cvpr}
\usepackage{times}
\usepackage{epsfig}
\usepackage{graphicx}
\usepackage{amsfonts} 
\usepackage{amsmath}
\usepackage{amssymb}
\usepackage{booktabs}       
\usepackage{color}
\usepackage{mathabx}

\newcommand{\Exp}{\mathbb{E}}
\newcommand{\QED}{{\footnotesize$\blacksquare$}}

\usepackage{amsthm} 

\newtheoremstyle{exampstyle}
  {1pt} 
  {0pt} 
  {\it } 
  {} 
  {\bfseries} 
  {.} 
  {.5em} 
  {} 
\theoremstyle{exampstyle} \newtheorem{lemma}{Lemma}
\theoremstyle{exampstyle} \newtheorem{remark}{Remark}

\usepackage[pagebackref=true,breaklinks=true,letterpaper=true,colorlinks,bookmarks=false]{hyperref}

\cvprfinalcopy 

\DeclareFontFamily{OT1}{pzc}{}
\DeclareFontShape{OT1}{pzc}{m}{it}{<-> s * [1.200] pzcmi7t}{}
\DeclareMathAlphabet{\mathpzc}{OT1}{pzc}{m}{it}
\newcommand{\pA}{\mathpzc{A}}
\newcommand{\pa}{\mathpzc{a}}
\newcommand{\pB}{\mathpzc{B}}
\newcommand{\pb}{\mathpzc{b}}
\newcommand{\pM}{\mathpzc{M}}
\newcommand{\pN}{\mathpzc{N}}
\newcommand{\pD}{\mathpzc{D}}
\newcommand{\pG}{\mathpzc{G}}
\newcommand{\pV}{\mathpzc{V}}
\newcommand{\pU}{\mathpzc{U}}
\newcommand{\pR}{\mathpzc{R}}

\begin{document}

\setlength{\textfloatsep}{5pt plus 1pt minus 2pt}

\title{Kernel-Based Training of Generative Networks}

\author{Kalliopi Basioti, George V.~Moustakides\\
Computer Science\\ 
Rutgers University, USA\\
{\tt\small (kib21@scarletmail.~gm463@)rutgers.edu}
\and
Emmanouil Z.~Psarakis\\
Computer Engineering and Informatics\\ 
University of Patras, Greece\\
{\tt\small psarakis@ceid.upatras.gr}
}

\maketitle
\thispagestyle{empty}

\begin{abstract}
Generative adversarial networks (GANs) are designed with the help of min-max optimization problems that are solved with stochastic gradient-type algorithms which are known to be non-robust. In this work we revisit a non-adversarial method based on kernels which relies on a pure minimization problem and propose a simple stochastic gradient algorithm for the computation of its solution. Using simplified tools from Stochastic Approximation theory we demonstrate that batch versions of the algorithm or smoothing of the gradient do not improve convergence. These observations allow for the development of a training algorithm that enjoys reduced computational complexity and increased robustness while exhibiting similar synthesis characteristics as classical GANs. 
\end{abstract}

\section{Background} 
 Since their first appearance \cite{Arjovsky,Goodfellow}, GANs have gained considerable attention and popularity, mainly due to their remarkable capability to produce, after proper training, synthetic data (usually images) that are realistically close to the data contained in their training set. The main challenge in designing GANs comes from the fact that their training algorithms require heavy computations that are primarily implementable on computationally powerful platforms. Such high computational needs arise not only because the size of the problems is usually large but also because the design of GANs requires the solution of min-max optimization problems. Stochastic gradient type algorithms employed for such cases very often exhibit non-robust behavior and slow rate of convergence, thus raising the computational needs considerably \cite{Creswell,Mescheder}.

In this work we focus, primarily, on the computational aspects of the training phase. Our intention is to develop a training algorithm which is simple and requires significantly less computations as compared to the current methods proposed in the literature and used in practice. In particular we will demonstrate, theoretically, that certain ideas as batch processing \cite{Masters2018RevisitingSB} and/or gradient smoothing \cite{Adam} that are used for the solution of min-max problems have, in fact, absolutely no effect in the proposed formulation and, therefore, can be ignored. These conclusions will help us shape our algorithmic scheme and suggest a simple and efficient form. In order to be able to develop our algorithm we need to recall certain key notions from the theory behind GANs and understand how it differs from the alternative approach we adopt here.

\begin{figure}[h!]
\centerline{\includegraphics[width=5cm]{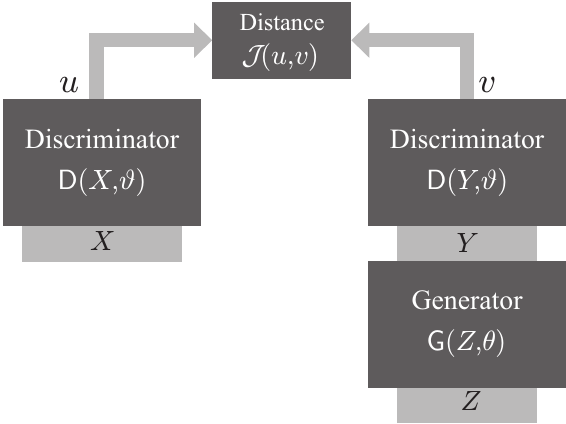}}
\caption{Representation of GAN architecture.}
\label{fig:1}
\end{figure}
Figure\,\ref{fig:1} captures the architecture employed during the training phase of GANs. There is a random vector $X$ with \textit{unknown} probability density function (pdf) $\mathsf{f}(X)$, with $X$ playing the role of a ``prototype'' random vector. The goal is to design a data-synthesis mechanism that generates realizations for the random vector $X$. For this goal we employ a nonlinear transformation $\mathsf{G}(Z,\theta)$, known as the \textit{Generator}, that transforms a random vector $Z$ of \textit{known} pdf (e.g.~Gaussian or Uniform) into a random vector $Y$. We would like to design the parameters $\theta$ of the transformation so that $Y$ is distributed according to $\mathsf{f}(\cdot)$. Under general assumptions such a transformation always exists \cite{Andrews,Box} and it can be efficiently approximated \cite{Cybenko} by a sufficiently large neural network, with $\theta$ summarizing the network parameters.

Adversarial approaches in order to make the proper selection of $\theta$ employ a \textit{second nonlinear transformation} $\mathsf{D}(\cdot,\vartheta)$ that transforms $X$ and $Y$ into suitable \textit{scalar} statistics $u=\mathsf{D}(X,\vartheta)$ and $v=\mathsf{D}(Y,\vartheta)$ and then compute a ``mismatch'' measure (not necessarily a distance) $\mathcal{J}(u,v)$ between the two random scalar quantities $u,v$. The second transformation $\mathsf{D}(\cdot,\vartheta)$ is also implemented with the help of a neural network, known as the \textit{Discriminator}. We are interested in the average mismatch between $u,v$ namely $\Exp_{u,v}[\mathcal{J}(u,v)]$ which, after substitution, can be written as
\begin{equation}
J(\theta,\vartheta)=\Exp_{X,Y}\big[\mathcal{J}\big(\mathsf{D}(X,\vartheta),\mathsf{D}(Y,\vartheta)\big)\big]=
\Exp_{X,Z}\left[\mathcal{J}\Big(\mathsf{D}(X,\vartheta),\mathsf{D}\big(\mathsf{G}(Z,\theta),\vartheta\big)\Big)\right].
\label{eq:1}
\end{equation}
For every selection of the generator parameters $\theta$ we would like to select the discriminator parameters $\vartheta$ so that the average mismatch between $u,v$ is maximized. In other words we design the discriminator to differentiate between the synthetic random vector $Y$ and the prototype random vector $X$, as much as possible. This worst-case performance we then attempt to minimize by selecting properly the generator parameters $\theta$. This leads to the following min-max optimization problem
\begin{equation}
\inf_{\theta}\,\sup_{\vartheta}\,J(\theta,\vartheta)
=\inf_{\theta}\,\sup_{\vartheta}\,\Exp_{X,Z}\left[\mathcal{J}\Big(\mathsf{D}(X,\vartheta),\mathsf{D}\big(\mathsf{G}(Z,\theta),\vartheta\big)\Big)\right].
\label{eq:min-max}
\end{equation}
Common selections for the mismatch function $\mathcal{J}(u,v)$ are:
\begin{itemize} \itemsep0em
\item $\mathcal{J}(u,v)=\log(u)+\log(1-v)$, $u,v\in[0,1]$, see \cite{Goodfellow}.

\item $\mathcal{J}(u,v)=\log(u)-\log(v)$, see \cite{Arjovsky}. 

\item $\mathcal{J}(u,v)=u-v$, see \cite{Arjovsky}.

\end{itemize}
It is clear that the generator generates realizations of the random vector $Y$ by transforming the realizations of $Z$. But how can we be assured that these realizations have the correct pdf namely $\mathsf{f}(\cdot)$?. To see that this is indeed the case we need to consider the generator and discriminator transformations $\mathsf{G}(\cdot),\mathsf{D}(\cdot)$ as being general functions not limited to the specific classes induced by the two neural networks. This immediately implies that by properly selecting $\mathsf{G}(\cdot)$ we can shape the pdf $\mathsf{g}(\cdot)$ of $Y$ into any pdf we desire \cite{Andrews}. In this idealized situation optimizing over $\theta$ amounts to optimizing over $\mathsf{G}(\cdot)$ and therefore over $\mathsf{g}(\cdot)$ and, similarly, optimizing over $\vartheta$ amounts to optimization over $\mathsf{D}(\cdot)$. Consequently, we can redefine the min-max optimization problem in \eqref{eq:min-max} under the following idealized frame
\begin{equation}
\inf_{\mathsf{g}}\,\sup_{\mathsf{D}}\,J(\mathsf{g},\mathsf{D})
=\inf_{\mathsf{g}}\,\sup_{\mathsf{D}}\, \iint\mathcal{J}\big(\mathsf{D}(X),\mathsf{D}(Y)\big) \mathsf{f}(X)\mathsf{g}(Y)dXdY,
\label{eq:ideal}
\end{equation}
where $\mathsf{D}(\cdot)$ any scalar valued nonlinear transformation, and $\mathsf{g}(\cdot)$ any pdf.
The min-max problems corresponding to the three examples of $\mathcal{J}(u,v)$ we mentioned before accept analytic solutions. In particular in the first case, for fixed $\mathsf{f}(\cdot),\mathsf{g}(\cdot)$ maximization over $\mathsf{D}(\cdot)$ is attained for $\mathsf{D}(\cdot)=\frac{\mathsf{f}(\cdot)}{\mathsf{f}(\cdot)+\mathsf{g}(\cdot)}$ and the resulting functional is minimized over $\mathsf{g}(\cdot)$ when $\mathsf{g}(\cdot)=\mathsf{f}(\cdot)$. In the second case, assuming that $|\mathsf{D}(\cdot)|\leq M$, maximization over $\mathsf{D}(\cdot)$ is achieved when $\mathsf{D}(\cdot)=e^{M\text{sgn}(\mathsf{f}(\cdot)-\mathsf{g}(\cdot))}$ and minimization over $\mathsf{g}(\cdot)$ yields, again, $\mathsf{g}(\cdot)=\mathsf{f}(\cdot)$. Similarly for the third case, maximization over $\mathsf{D}(\cdot)$ is achieved for $\mathsf{D}(\cdot)=M\text{sgn}(\mathsf{f}(\cdot)-\mathsf{g}(\cdot))$ and minimization over $\mathsf{g}(\cdot)$ when $\mathsf{g}(\cdot)=\mathsf{f}(\cdot)$. As we can see all three min-max problems result in different optimum discriminator functions but agree in the final solution for $\mathsf{g}(\cdot)$, namely $Y$ is shaped to have the same pdf $\mathsf{f}(\cdot)$ as the prototype random vector $X$. 

Since in the original min-max problem \eqref{eq:min-max} we limit the two transformations to be within the two classes induced by the input/output relationship of the corresponding neural network, it is clear that \eqref{eq:min-max} constitutes an approximation to the ideal setup captured by \eqref{eq:ideal}. This implies that the output $Y$ of the generator will follow a pdf $\mathsf{g}(\cdot)$ which will be an \textit{approximation} to the desired pdf $\mathsf{f}(\cdot)$ of the prototype random vector $X$. This approximation not only depends on the richness of the transformation class induced by the generator structure but, also, on the corresponding richness of the discriminator structure. As long as one of the two structures does not approximate sufficiently close the corresponding ideal transformation (when for example the neural network does not have sufficient number of layers), \textit{the design will fail} in the sense that the realizations of $Y$ will not follow the desired pdf $\mathsf{f}(\cdot)$ of the prototype $X$. The min-max optimization problem becomes more challenging because, as we mentioned, the pdf of $X$ is unknown and, instead, we are given a collection $\{X_1,\ldots,X_N\}$ of independent realizations of $X$ (the training set) drawn from $\mathsf{f}(\cdot)$. 
\vskip0.1cm
\begin{remark}\label{rem:1}
Even though the goal is to design a generator network, with GANs we simultaneously require the design of an additional neural network, the discriminator. This requirement increases the number of parameters to be estimated considerably and, consequently, the computational complexity.
\end{remark}
\noindent Furthermore the algorithmic solution of \eqref{eq:min-max} relies on alternating stochastic gradient-type algorithms and the presence of two antagonistic optimization problems translates into an increased number of updates in the implementation which are also known to be \textit{non-robust} \cite{Bengio-tricks-arxiv2012,Creswell,Mescheder}.

\section{A non-adversarial approach}
 Let us now see how we can accomplish a similar approximation for the output pdf of the generator without the need of a discriminator. We are going to revisit the idea suggested in \cite{Dziugaite,Gretton:2012:KTT:2503308.2188410} which is based on kernel functions. Let $U,V$ be vectors of the same dimension of $X$ and consider a scalar function $\mathsf{k}(U,V)$ which is symmetric, i.e. $\mathsf{k}(U,V)=\mathsf{k}(V,U)$ and positive definite, namely, for every scalar function $\phi(\cdot)$ it satisfies
$$
\iint \phi(U)\mathsf{k}(U,V)\phi(V)\,dU\,dV\geq0,
$$
with equality to 0 if and only if $\phi(\cdot)=0$.

For two pdfs $\mathsf{f}(\cdot),\mathsf{g}(\cdot)$ with $\mathsf{f}(\cdot)$ fixed and $\mathsf{g}(\cdot)$ to be determined, we define a distance measure as a function of $\mathsf{g}(\cdot)$ as follows
\begin{equation}
J(\mathsf{g})=
\iint\big(\mathsf{f}(U)-\mathsf{g}(U)\big)\mathsf{k}(U,V)\big(\mathsf{f}(V)-\mathsf{g}(V)\big)\,dU\,dV.
\label{eq:1.1}
\end{equation}
An immediate consequence of the positive definiteness property of the kernel is that the solution to the
minimization problem $\inf_{\mathsf{g}}J(\mathsf{g})$ is, obviously, $\mathsf{g}(\cdot)=\mathsf{f}(\cdot)$. Let us now write the same distance using expectations. If $X$, $Y^1$, $Y^2$ are \textit{independent} random vectors with $X$ following $\mathsf{f}(\cdot)$ and $Y^1,Y^2$ following $\mathsf{g}(\cdot)$ then, the double integral in \eqref{eq:1.1} can be expressed as
\begin{multline}
J(\mathsf{g})=c-\Exp_{X,Y^1}[\mathsf{k}(Y^1,X)]
-\Exp_{X,Y^2}[\mathsf{k}(X,Y^2)]+\Exp_{Y^1,Y^2}[\mathsf{k}(Y^1,Y^2)]=\\
c+\Exp_{X,Y^1\!\!,Y^2}[\mathsf{k}(Y^1,Y^2)-\mathsf{k}(Y^1,X)-\mathsf{k}(X,Y^2)].
\label{eq:3}
\end{multline}
where $c=\iint \mathsf{f}(U)\mathsf{k}(U,V)\mathsf{f}(V)dU\,dV$ is constant, not related to $\mathsf{g}(\cdot)$. Eq.\,\eqref{eq:3} is simply an alternative way to rewrite the metric introduced in \eqref{eq:1.1}, consequently its minimization with respect to $\mathsf{g}(\cdot)$ still results in the desired equality $\mathsf{g}(\cdot)=\mathsf{f}(\cdot)$.

The next step consists in abandoning the ideal world expressed by \eqref{eq:3}. If $Y$ is the output $Y=\mathsf{G}(Z,\theta)$ of the generator, this suggests that $Y^1,Y^2$ correspond to inputs $Z^1,Z^2$. The two random input vectors must be statistically independent in order for the same property to be inherited by the two outputs $Y^1,Y^2$. From \eqref{eq:3}, by substituting $Y^i=\mathsf{G}(Z^i,\theta),~i=1,2$ and using the symmetry of the kernel, we can define an average distance as a function of the generator parameters $\theta$ as follows\footnote{We prefer the \textit{symmetric} form in \eqref{eq:f_problem} instead of the expectation of $\mathsf{k}(\mathsf{G}(Z^1,\theta),\mathsf{G}(Z^2,\theta))-2\mathsf{k}(\mathsf{G}(Z^1,\theta),X)$ adopted in \cite{Dziugaite}.}
\begin{equation}
J(\theta)
=\Exp_{X,Z^1\!\!,Z^2}\big[
\mathsf{k}\big(\mathsf{G}(Z^1,\theta),\mathsf{G}(Z^2,\theta)\big)
-\mathsf{k}\big(\mathsf{G}(Z^1,\theta),X\big)-\mathsf{k}\big(\mathsf{G}(Z^2,\theta),X\big)
\big],
\label{eq:f_problem}
\end{equation}
where in the proposed measure we left out the constant term $c$ since it does not depend on $\theta$. Performing the minimization $\inf_{\theta}J(\theta)$ generates a neural network whose output $Y$ will have a pdf that approximates the desired pdf $\mathsf{f}(\cdot)$ in the sense of the average distance we introduced in \eqref{eq:1.1}. This is clearly the equivalent of the min-max problem in \eqref{eq:min-max} and, as we can see, it involves a \textit{pure minimization}.

\section{Properties of training algorithms}
Because the problem we are concerned with involves only minimization, this allows for the use of classical stochastic gradient algorithms. What is also appealing, is that we have a rich arsenal of theoretical results coming from Stochastic Approximation Theory \cite{Benveniste} that can support the training algorithm we intend to propose. In fact our goal is to arrive at an algorithmic scheme that has reduced computational complexity and demonstrate, theoretically and/or with simulations, that there is no significant performance loss in doing so. Actually the relevant properties we are going to use in order to propose our algorithmic scheme will be presented under a more general frame not limited to the specific optimization problem defined in the previous section.

\subsection{Stochastic approximation}\label{ssec:3.1}
\begin{figure}[h!]
\centerline{\includegraphics[width=7.0cm]{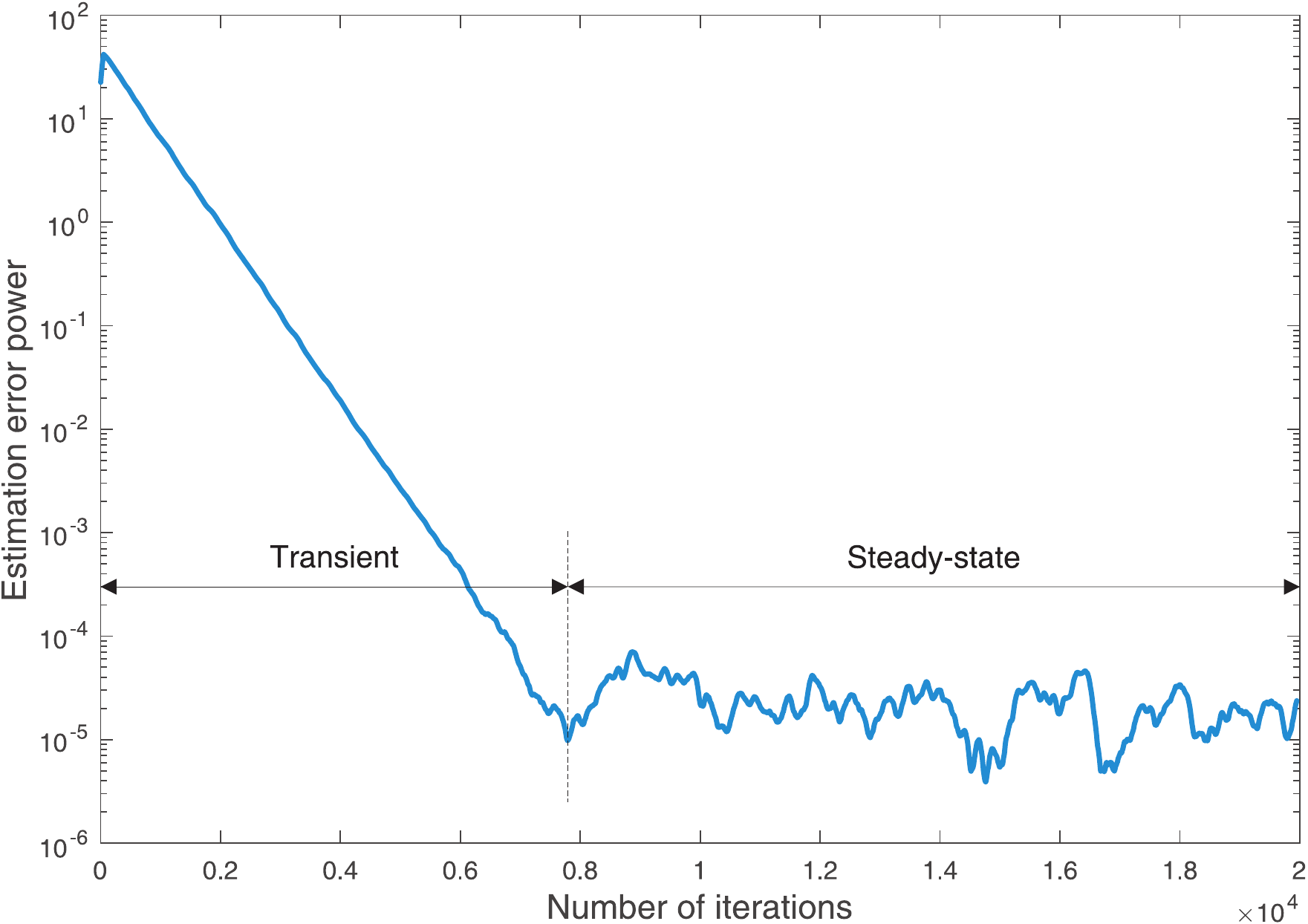}}
\caption{Typical form of estimation error power as a function of number of iterations for a linear regression model.}
\label{fig:2}
\end{figure}

 Suppose we are given a function $\mathsf{h}(W,\theta)$ where $W$ denotes a random vector for which we have available a sequence of independent realizations $\{W_t\}$. Consider the following optimization problem
\begin{equation}
\inf_\theta\Exp_W[\mathsf{h}(W,\theta)]
\label{eq:min_h}
\end{equation}
It is then well known that the stochastic gradient algorithm 
\begin{equation}
\theta_t=\theta_{t-1}-\mu \mathsf{H}(W_t,\theta_{t-1}),~~\mathsf{H}(W,\theta)=\nabla_\theta\mathsf{h}(W,\theta) 
\label{eq:algorithm}
\end{equation}
where $\mu$ denotes the learning rate of the algorithm, can lead to a (local) minimizer of \eqref{eq:min_h} without knowing the probability distribution of $W$. The algorithm in \eqref{eq:algorithm} can be characterized \cite{Benveniste} by the average trajectory $\bar{\theta}_t$ and the corresponding random perturbations $v_t$ as $\theta_t=\bar{\theta}_t+v_t$ where
$$
\bar{\theta}_t=\bar{\theta}_{t-1}-\mu\Exp_W[\mathsf{H}(W,\bar{\theta}_{t-1})],
$$ 
while for $v_t=\theta_t-\bar{\theta}_t$ we have the steady-state description $\Exp[v_\infty v_\infty^\intercal]=\lim_{t\to\infty}\Exp[v_tv_t^\intercal]=\mu\big(1+o(1)\big)Q$ with the matrix $Q$ satisfying the following Lyapunov equation
\begin{equation}
CQ+QC^\intercal=\Exp_W[\mathsf{H}(W,\theta_*)\mathsf{H}^\intercal(W,\theta_*)].
\label{eq:Liapunov}
\end{equation}
Vector $\theta_*$ is the true (local) minimizer of \eqref{eq:min_h} 
and $C$ is the Hessian of $\Exp_W[\mathsf{H}(W,\theta)]$ evaluated at $\theta_*$. As it is explained in \cite{Benveniste}, the mean trajectory captures the transient phase while the perturbation part becomes leading during the steady-state of the corresponding algorithmic run. This is also graphically depicted in Figure\,\ref{fig:2} 
 for the case of a simple regression model of the form $y_t=\theta_*^\intercal X_t+w_t$, where $X_t$ has length 5, $\{X_t\}$ is i.i.d.~zero-mean Gaussian with unit covariance matrix and $\{w_t\}$ is i.i.d.~additive zero-mean Gaussian noise with variance $0.1$ and independent from $\{X_t\}$. We are interested in $\inf_\theta\Exp_{y,X}[(y-\theta^\intercal X)^2]$. In the stochastic gradient descent version $\theta_t=\theta_{t-1}+\mu(y_t-\theta_{t-1}^\intercal X_t)X_t$ we select the learning rate $\mu=0.001$.
The average trajectory and the steady-state performance introduced before can be used as a means to compare algorithms.
\vskip0.1cm
\begin{remark}\label{rem:2}
When two algorithms have similar average trajectories and exhibit the same steady-state behavior, they are practically equivalent in performance.
\end{remark}
\vskip0.1cm
\noindent This simple rule which, of course, makes sense will allow us to examine whether certain alternative versions of the classical algorithm in \eqref{eq:algorithm} can indeed improve its convergence characteristics.

\subsection{Does batch processing improve convergence?}\label{ssec:batch}
A widespread impression \cite{Dziugaite,Masters2018RevisitingSB} is that if we use data in batches $\{W_{(n-1)K+1},\ldots,W_{nK}\}$ of length $K$ and approximate $\Exp_W[\mathsf{H}(W,\theta)]$ with $\frac{1}{K}\sum_{j=0}^{K-1}\mathsf{H}(W_{nK-j},\theta)$ instead of $\mathsf{H}(W_t,\theta)$, then the corresponding algorithm
\begin{equation}
\theta_n'=
\theta_{n-1}'-\frac{\mu'}{K}\sum_{j=0}^{K-1}\mathsf{H}(W_{nK-j},\theta_{n-1}')
\label{eq:batch}
\end{equation}
converges faster than \eqref{eq:algorithm}. Considering speed in terms of iterations is actually completely unfair since each iteration in \eqref{eq:batch} involves the usage of $K$ vectors $W_j$ and $K$ gradient computations instead of a single $W_j$ and a single gradient computation in the classical scheme \eqref{eq:algorithm}. In order for the comparison to be correct we need to count speed in terms of the number of $W_j$ vectors already used or the number of gradient computations already performed. Following this principle, \eqref{eq:batch} should be expressed as
\begin{equation}
\theta_{nK}'=
\theta_{(n-1)K}'-\frac{\mu'}{K}\sum_{j=0}^{K-1}\mathsf{H}(W_{nK-j},\theta_{(n-1)K}').
\label{eq:batch2}
\end{equation}

Returning to the classical version, from Stochastic Approximation theory we know that the algorithm in \eqref{eq:algorithm} has a natural ability for averaging/smoothing. This can become apparent if we subsample \eqref{eq:algorithm} every $K$ iterations and expand the formula across $K$ consecutive updates
\begin{equation}
\theta_{nK}=
\theta_{(n-1)K}-\mu\sum_{j=0}^{K-1}\mathsf{H}(W_{nK-j},\theta_{nK-j-1}).
\label{eq:algorithm2}
\end{equation}
As we can see by selecting $\mu=\mu'/K$, \eqref{eq:batch2} and \eqref{eq:algorithm2} become very similar. Of course we observe that in the latter the parameter estimates are different in each term of the sum as opposed to the former where these estimates are all the same. We should however note that, since $\mu$ is very small, $\theta_t$ changes very slowly resulting in minor differences between $\theta_{nK-j-1}$ and $\theta_{(n-1)K}$. Following Remark\,\ref{rem:2}, we can make a formal claim by computing the average trajectories and the steady-state perturbation covariance matrices of the two versions. The following lemma compares the two algorithms.
\begin{lemma}\label{lem:1}
The average trajectories in \eqref{eq:batch2} and \eqref{eq:algorithm2} are given respectively by
\begin{align*}
\bar{\theta}_{nK}'&=\bar{\theta}_{(n-1)K}'-\mu'\Exp_W[\mathsf{H}(W,\bar{\theta}_{(n-1)K}')]\\
\bar{\theta}_{nK}&=\bar{\theta}_{(n-1)K}-\mu'\big(1+O(\mu')\big)\Exp_W[\mathsf{H}(W,\bar{\theta}_{(n-1)K})],
\end{align*}
while the steady-state perturbation covariance matrix in both cases satisfies $\Exp[v_\infty v_\infty^\intercal]=\Exp[v_\infty'(v_\infty')^\intercal]=\frac{\mu'}{K}\big(1+o(1)\big)Q$ and $Q$ is the solution of \eqref{eq:Liapunov}. 
\end{lemma}
\vskip0.1cm
\noindent\textbf{Proof:}~The first equation is a direct consequence of the definition of the average trajectory. For the second we assume sufficient smoothness of the vector function $\Exp_W[\mathsf{H}(W,\theta)]$ and apply a Taylor expansion around $\bar{\theta}_{(n-1)K}$ after expressing $\bar{\theta}_{nK-j-1}=\bar{\theta}_{(n-1)K}+O(\mu')$. Finally, the computation of the perturbation covariance matrices is also straightforward and since it involves the exact minimizer $\theta_*$ it is the same for both algorithms.\,\QED

\begin{figure}[h!]
\centerline{\includegraphics[width=7.5cm]{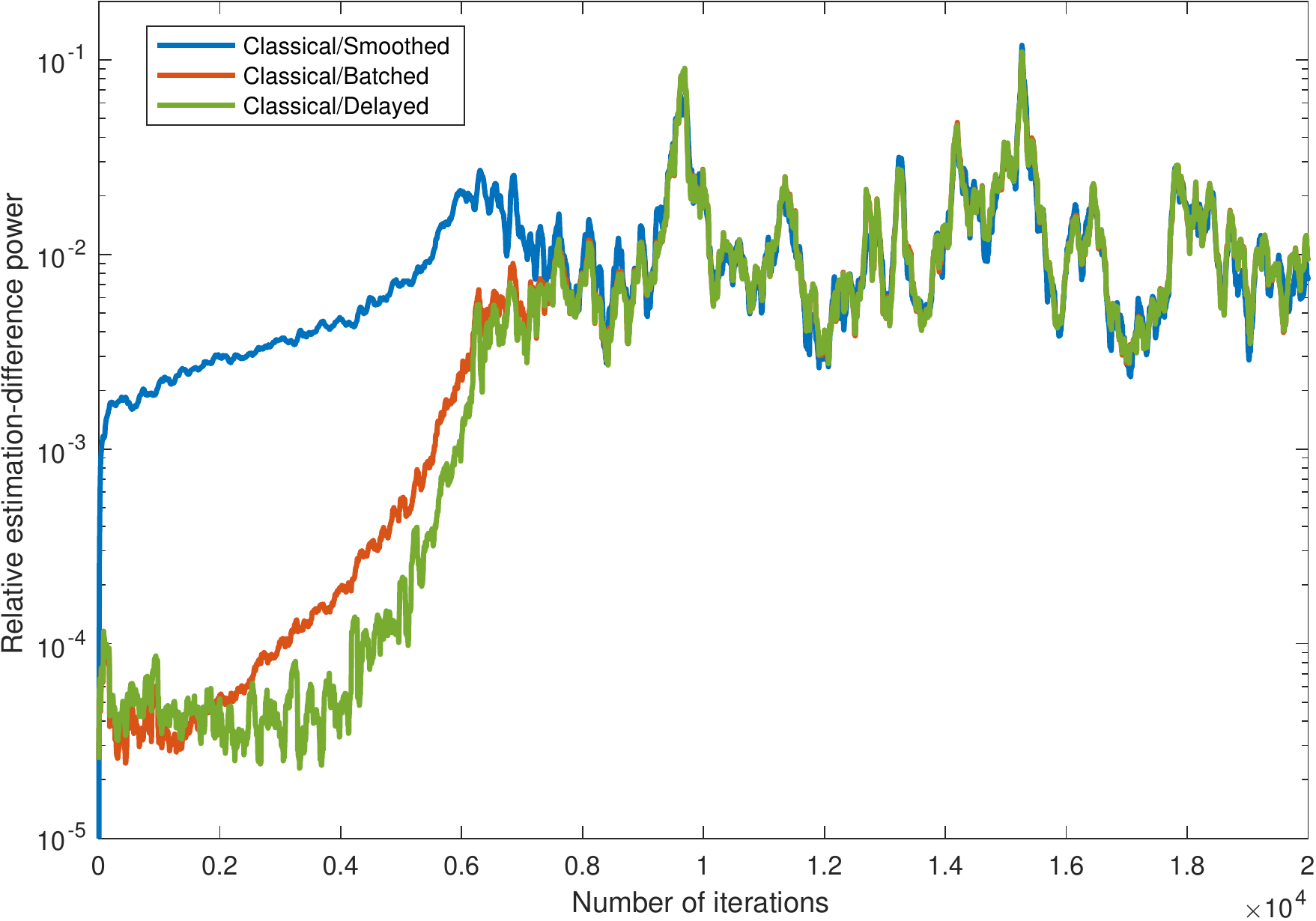}}
\caption{Relative estimation-difference power between classical and batched for $K=10$ (red); classical and smoothed for $\rho=0.9$ (blue); and classical and delayed for $k=5$ (green).}
\label{fig:3}
\end{figure}

Lemma\,\ref{lem:1} implies that batching has actually no noticeable effect during initial convergence and during steady-state.~This is also confirmed from Figure\,\ref{fig:3}
where we present the relative estimation-difference power $2\|\theta_{t}-\theta_{t}'\|^2/(\|\theta_{t}-\theta_*\|^2+\|\theta_{t}'-\theta_*\|^2)$ (red) of the two algorithms, again for the case of the simple regression model. As we can see this quantity is very small during the transient phase while during steady-state it becomes proportional to $\mu'$. The latter can be verified by performing simulations with different $\mu'$ values and observing the corresponding change in the relative power during steady-state.

Actually, Figure\,\ref{fig:3} allows us to make a claim that is far stronger than Lemma\,\ref{lem:1}: Not only the two versions exhibit similar first and second order moments over iterations (i.e.~similar average trajectories and steady-state perturbation covariance matrices, as stated in Lemma\,\ref{lem:1}), but their \textit{actual} estimates $\theta_t,\theta_t'$ are very close to each other, provided of course that the two algorithms use the same data, synchronously.
\vskip0.1cm
\begin{remark}\label{rem:3}
Even though no convergence speed improvement is observed, batch processing can be beneficial since it can exploit existing parallel or vectorized processing capabilities of the computational platform. 
\end{remark}
\vskip0.1cm
\noindent Indeed, as we demonstrated, per gradient computation there is no improvement in convergence, however, if there are parallel processing units (or vectorized computational capabilities) we can perform multiple gradient computations simultaneously and reduce the overall physical computational time \cite{Bengio-tricks-arxiv2012}.

\subsection{Does smoothing improve convergence?}\label{ssec:smooth}
 Another popular variation of the classical stochastic gradient algorithm consists in replacing the instantaneous gradient with a smoothed version updated over each iteration. In particular, instead of \eqref{eq:algorithm}, in \cite{Adam} it is proposed as alternative
\begin{align}
\tilde{\mathsf{H}}_t&=\rho\tilde{\mathsf{H}}_{t-1}+(1-\rho)\mathsf{H}(W_t,\theta_{t-1}'') \label{eq:smoothH}\\
\theta_t''&=\theta_{t-1}''-\mu \tilde{\mathsf{H}}_t \label{eq:algorithm3}.
\end{align}
The smoothing in \eqref{eq:smoothH} corresponds to an exponential windowing in place of the orthogonal window employed in the batch implementation. This can be seen from the expansion 
$$
\tilde{\mathsf{H}}_t=
(1-\rho)\sum_{j=0}^{t-1}\rho^{j}\mathsf{H}(W_{t-j},\theta_{t-j-1}'').
$$
A typical value of $\rho$ is 0.9 which implies that very quickly the contribution of past gradients in the sum, due to the term $\rho^j$, becomes negligible and the sum appears as having practically a fixed number of terms. In fact it is commonly considered in signal processing that an exponential window has the same effect as an orthogonal window of length $K=\frac{1}{1-\rho}$. This makes smoothing similar to batch processing and, therefore, it is expected not to provide any noticeable difference compared to the original algorithm \eqref{eq:algorithm}. 

We could offer a formal proof to our previous argument by finding, as before, the average trajectory and the steady-state covariance matrix and show that they are similar to the original version. Instead, for simplicity we provide a simulation example in the hope that it is equally convincing. In Figure\,\ref{fig:3} we plot the relative estimation-difference power $2\|\theta_{t}-\theta_{t}''\|^2/(\|\theta_{t}-\theta_*\|^2+\|\theta_{t}''-\theta_*\|^2)$ (blue) for the regression model example where in \eqref{eq:smoothH} we used $\rho=0.9$. As we observe, again the two algorithms provide similar estimates with the relative estimation-difference power being of the order of $\mu'$ during steady-state and much smaller during the transient phase. 

\subsection{Does gradient computation using past parameter estimates affect convergence?}\label{ssec:delay}
Next we would like to examine the effect on the algorithmic performance when in \eqref{eq:algorithm} the computation of the gradient is performed not by using $\theta_{t-1}$ but $\theta_{t-k},~k>1$. In other words we consider the algorithm
\begin{equation}
\theta_t'''=\theta_{t-1}'''-\mu \mathsf{H}(W_t,\theta_{t-k}''').
\label{eq:algorithm4}
\end{equation}
Again, computing the average trajectory and the steady-state perturbation covariance matrix we can show that the two algorithms in \eqref{eq:algorithm} and \eqref{eq:algorithm4} are described by similar equations. In particular for \eqref{eq:algorithm4} we need to use that fact that $\bar{\theta}_{t-k}'''=\bar{\theta}_{t-1}'''+O(\mu)$ which will inflict an $O(\mu^2)$ difference as compared to the average trajectory $\bar{\theta}_t$ of \eqref{eq:algorithm}. The steady-state behavior on the other hand will be the same. In Figure\,\ref{fig:3}, as before, we plot $2\|\theta_{t}-\theta_{t}'''\|^2/(\|\theta_{t}-\theta_*\|^2+\|\theta_{t}'''-\theta_*\|^2)$ (green) for a delay $k=5$. As we can see, if for the computation of the gradient we use a delayed version of our parameter estimate, this has only a negligible effect on the overall convergence behavior of the algorithm. 
\vskip0.1cm
\begin{remark}\label{rem:4}
The previous properties apply to every algorithm in the form of \eqref{eq:algorithm}. We should however emphasize that the computational schemes employed in classical GANs for solving \eqref{eq:min-max} do not fall under this frame.
\end{remark}
\vskip0.1cm
\noindent Indeed for min-max problems each update of $\theta$ (generator parameters) is followed by several updates of $\vartheta$ (discriminator parameters).~Consequently batching/smooth\-ing/delay\-ing may affect these algorithms differently. As far as the class of algorithms captured by \eqref{eq:algorithm} is concerned, which are the focus of this work, we believe we have provided sufficient evidence that these modifications have no significant effect on the characteristics of the algorithm.

\section{Proposed algorithmic scheme}
 Let us now return to the problem of interest, namely the minimization of $J(\theta)$ which is defined in \eqref{eq:f_problem}. Because of the properties described in Sections\,\ref{ssec:batch}, \ref{ssec:smooth} it is clear that we will adopt a simple version without smoothing (which is common in adversarial approaches \cite{Adam}). The property mentioned in Section\,\ref{ssec:delay} will be used after we make the presentation of the first version of our algorithm and it will result in a significant computational reduction without any noticeable sacrifice in performance.
 
 Following \eqref{eq:min_h} and \eqref{eq:algorithm}, at each iteration $t$ we need to provide \textit{two} statistically independent realizations $Z_t^1,Z_t^2$ of the input vector $Z$ and one realization $X_t$ of $X$. As we pointed out $Z_t^1,Z_t^2$ can be generated since their pdf is assumed known (e.g.~Gaussian or Uniform) while $X_t$ is available from the training data set $\{X_1,\ldots,X_N\}$. 

We would like to point out that stochastic gradient type algorithms for the minimization of $J(\theta)$ were previously proposed in \cite{Dziugaite,Li}. In \cite{Li} expectation is replaced by averaging over the whole set of available training data. Consequently each iteration requires a considerable amount of gradient computations. In \cite{Dziugaite} this problem is reduced since they propose the use of small-sized (mini) batches. Specifically they define one batch with $X_i$'s and a second with pairs $(Z_j^1,Z_j^2)$. Because each $X_i$ from the first batch is combined with \textit{every pair} $(Z^1_j,Z^2_j)$ in the second batch, the number of gradient evaluations is still elevated. 

Consider now a neural network with two layers. In particular if $Z,Y$ are the input and output respectively, we define
$$
W=\pA Z+\pa,~~S=d(W);~~T=\pB S+\pb,~~Y=g(T)
$$
where $d(w),g(w)$ denote scalar nonlinearities with $d(W),g(W)$ meaning that $d(w),g(w)$ is applied to each element of the vector $W$. Matrices $\pA,\pB$ and vectors $\pa,\pb$ constitute the parameters of the two layers with each matrix forming the linear combination and the vector providing the offset. The input vector $Z$ is usually of much smaller dimension as compared to $X$ and $Y$ and we reach the target dimension progressively. For our algorithm we consider only fully connected neural networks without any special constraint on their coefficients.

To apply the stochastic gradient algorithm, the most crucial part is the computation of the gradient of the kernel $\mathsf{k}(Y,U)$ with respect to the parameter matrices and vectors. These parameters affect the kernel value through $Y$. Interestingly, there is a very simple recursive formula that allows for the computation of the corresponding derivatives. This is presented in the following lemma.
\vskip0.1cm
\begin{lemma}\label{lem:2}
Denote with $\nabla_{[\pA\,\pa]}\mathsf{k}(Y,U),\nabla_{[\pB\,\pb]}\mathsf{k}(Y,U)$ the gradients of the kernel $\mathsf{k}(Y,U)$ with respect to the elements of the matrices $[\pA\,\pa],[\pB\,\pb]$ that affect $Y$ and with the partial derivatives arranged into a matrix of the same dimensions. To compute the gradients define
$$
\pV=g'(T)\odot\nabla_Y\mathsf{k}(Y,U),~\pU=d'(W)\odot(\pB^\intercal V)
$$
with $d'(w),g'(w)$ denoting the derivatives of $d(w),g(w)$ and $A\odot B$ denoting the element-by-element multiplication of two matrices $A,B$ of the same dimensions. Then the two gradients take the simple form
\[
\nabla_{[\pA\,\pa]}\mathsf{k}(Y,U)=\pU[S^\intercal\,1],~~~
\nabla_{[\pB\,\pb]}\mathsf{k}(Y,U)=\pV[Z^\intercal\,1].
\]
\end{lemma}

\noindent\textbf{Proof:}~The proof of this lemma presents no particular difficulty. It only requires careful housekeeping of the various partial derivatives. It is also worth mentioning that the gradients of the kernel function with respect to the parameters of each layer turn out to form a rank-one matrix.\,\QED

Let us now limit ourselves to the Gaussian kernel case $\mathsf{k}(U,V)=e^{-\|U-V\|^2/h}$. We then have 
\[
\nabla_Y\mathsf{k}(Y,U)=-\frac{2}{h}e^{-\frac{1}{h}\|Y-U\|^2}(Y-U).
\]
The function that plays the role of $\mathsf{h}$ in \eqref{eq:min_h} is
\[
\mathsf{k}(Y^1,Y^2)-\mathsf{k}(Y^1,X)-\mathsf{k}(Y^2,X)
\]
for which we must compute the gradient with respect to the parameters of each layer. The first version of the algorithm is depicted in Table\,\ref{tab:1}. 
\begin{table}[t!]
\caption{Preliminary version of training algorithm.}
\label{tab:1}
\centering
\begin{tabular}{l}
\toprule
Initialize~$\pA_0,\pB_0$ using the method in \cite{Glorot} and set $\pa_0,\pb_0$ to zero.\\
\midrule
Available from iteration $t-1$:~$\pA_{t-1},\pB_{t-1},\pa_{t-1},\pb_{t-1}$.\\
\midrule
At iteration $t$:\\
Generate~inputs: $Z_t^1,Z_t^2$ and select~$X_t$ from training set; cycle data if exhausted.\\
Compute layer outputs for $i=1,2$:\\
~~~~$W_t^i=\pA_{t-1}Z_t^i+\pa_{t-1}$,~$S_t^i=d(W_t^i)$\\
\addlinespace[2pt]
~~~~$T_t^i=\pB_{t-1}S_t^i+\pb_{t-1}$,~$Y_t^i=g(T_t^i)$\\
\addlinespace[2pt]
Compute gradients:\\
~~~~$\pR^1_{\,t}=(Y_t^1-X_t)e^{-\frac{\|Y_t^1-X_t\|^2}{h}}-(Y_t^1-Y_t^2)e^{-\frac{\|Y_t^1-Y_t^2\|^2}{h}}$\\
~~~~$\pR_{\,t}^2=(Y_t^2-X_t)e^{-\frac{\|Y_t^2-X_t\|^2}{h}}-(Y_t^2-Y_t^1)e^{-\frac{\|Y_t^2-Y_t^1\|^2}{h}}$\\
\addlinespace[2pt]
~~~~For $i=1,2$ compute:\\
~~~~$\pV_t^i=g'(T_t^i)\odot \pR_{\,t}^i$\\
~~~~$\pU_t^i=d'(W_t^i)\odot (B_{t-1}^\intercal \pV_t^i)$\\
~~~~$\pG_t=\pV_t^1[(S_t^1)^\intercal\,1]+\pV_t^2[(S_t^2)^\intercal\,1]$\\
\addlinespace[2pt]
~~~~$\pD_t=\pU_t^1[(Z_t^1)^\intercal\,1]+\pU_t^2[(Z_t^2)^\intercal\,1]$\\
\addlinespace[2pt]
Update parameter estimates:\\
~~~~$[\pB_t\,\pb_t]=[\pB_{t-1}\,\pb_{t-1}]-\mu\pG_t$\\
\addlinespace[2pt]
~~~~$[\pA_t\,\pa_t]=[\pA_{t-1}\,\pa_{t-1}]-\mu\pD_t$\\
\midrule
Repeat until some stopping rule is satisfied.\\
\bottomrule
\addlinespace[0.1cm]
\end{tabular}
\end{table}

As we see from Table\,\ref{tab:1}, in each iteration $t$ we need to compute the layer outputs and the corresponding gradients for two different inputs $Z_t^1,Z_t^2$. It is exactly here that we intend to use the result of Section\,\ref{ssec:delay}. We propose to compute the layer outputs and gradients corresponding to  a \textit{single} input $Z_t$ while as second input we use the one generated during the previous iteration along with its outputs and gradients. This poses no problem since $Z_t$ and $Z_{t-1}$ are independent, as required by our analysis. Finally, a last issue we would like to address is normalization. In order for the learning rate to become data independent, it is necessary that the gradient, before being used in the update, to be normalized. For this reason we adopt the scheme proposed in \cite{Hinton} with $\lambda=0.999$. Incorporating all these points into our algorithmic structure produces the final result depicted in Table\,\ref{tab:2}. Regarding our notation, if $A$ is a matrix then $(A)_{ij}$ denotes its $ij$-th element.
\begin{table}[h!]
\vskip-0.55cm
\caption{Final version of training algorithm.}
\label{tab:2}
\centering
\begin{tabular}{l}
\toprule
Initialize~$\pA_0,\pB_0,$ using the method in \cite{Glorot} and set $\pa_0,\pb_0,Z_0,S_0,Y_0,\pV_0,\pU_0,$\\
~~~~$\pM_0,\pN_0$ to zero.\\
\addlinespace[-1pt]
\midrule
Available from iteration $t-1$:~$\pA_{t-1},\pB_{t-1},\pa_{t-1},\pb_{t-1},Z_{t-1},S_{t-1},Y_{t-1},\pV_{t-1},$\\
~~~~$\pU_{t-1},\pM_{t-1},\pN_{t-1}$.\\
\addlinespace[-1pt]
\midrule
\end{tabular}
\end{table}

\begin{table}[h!]
\centering
\begin{tabular}{l}
\midrule
At iteration $t$:\\
\addlinespace[1pt]
Generate~input: $Z_t$ and select~$X_t$ from training set; cycle data if exhausted.\\
\addlinespace[2pt]
Compute layer outputs:\\
~~~~$W_t=\pA_{t-1}Z_t+\pa_{t-1}$,~$S_t=d(W_t)$\\
~~~~$T_t=\pB_{t-1}S_t+\pb_{t-1}$,~$Y_t=g(T_t)$\\
\addlinespace[2pt]
Compute gradients:\\
~~~~$\pR_{\,t}=(Y_t-X_t)e^{-\frac{\|Y_t-X_t\|^2}{h}}-(Y_t-Y_{t-1})e^{-\frac{\|Y_t-Y_{t-1}\|^2}{h}}$\\
~~~~$\pV_t=g'(T_t)\odot \pR_{\,t}$\\
~~~~$\pU_t=d'(W_t)\odot (B_{t-1}^\intercal \pV_t)$\\
~~~~$\pG_t=\pV_t[S_t^\intercal\,1]+\pV_{t-1}[S_{t-1}^\intercal\,1]$\\
~~~~$\pD_t=\pU_t[Z_t^\intercal\,1]+\pU_{t-1}[Z_{t-1}^\intercal\,1]$\\
\addlinespace[2pt]
Estimate power of gradient elements:\\
~~~~$(\pM_t)_{ij}=\lambda (\pM_{t-1})_{ij}+(1-\lambda)(\pG_t)_{ij}^2$\\
~~~~$(\pN_t)_{ij}=\lambda (\pN_{t-1})_{ij}+(1-\lambda)(\pD_t)_{ij}^2$\\
\addlinespace[2pt]
Update parameter estimates:\\
~~~~$([\pB_t\,\pb_t])_{ij}=([\pB_{t-1}\,\pb_{t-1}])_{ij}-\mu\frac{(\pG_t)_{ij}}{\sqrt{(\pM_t)_{ij}}}$\\
~~~~$([\pA_t\,\pa_t])_{ij}=([\pA_{t-1}\,\pa_{t-1}])_{ij}-\mu\frac{(\pD_t)_{ij}}{\sqrt{(\pN_t)_{ij}}}$\\
\midrule
Repeat until stopping some rule is satisfied.\\
\hphantom{Available from iteration $t-1$:~$\pA_{t-1},\pB_{t-1},\pa_{t-1},\pb_{t-1},Z_{t-1},S_{t-1},Y_{t-1},\pV_{t-1},$}\\
\addlinespace[-15pt]
\bottomrule
\addlinespace[-15pt]
\end{tabular}
\end{table}

\section{Experiments}
\begin{figure}[b!]
\centerline{\includegraphics[width=7.0cm]{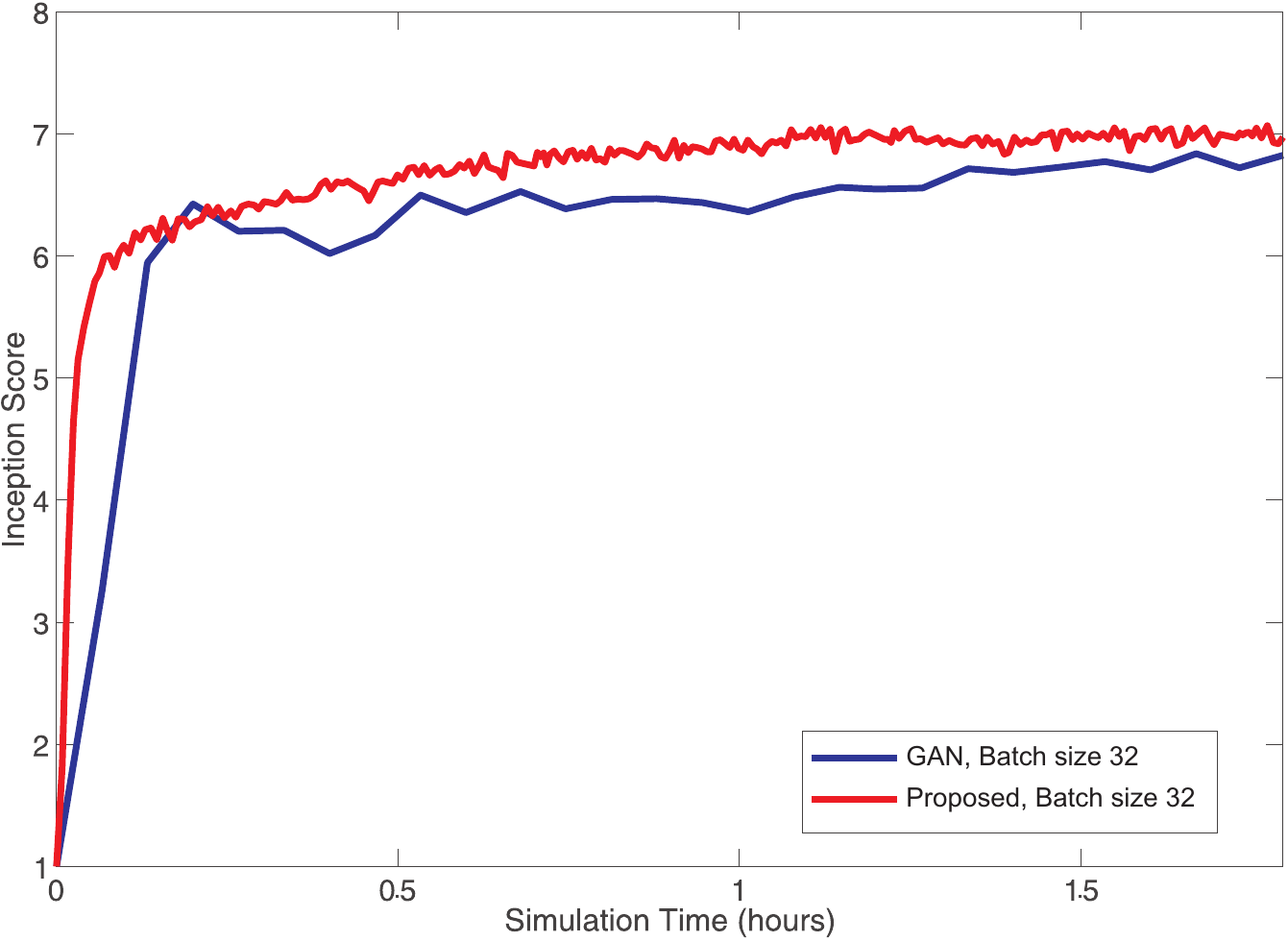}}
\caption{Inception score for proposed and GAN \cite{DBLP:journals/corr/abs-1802-03446,huang2018an} for the MNIST dataset.}
\label{fig:fig5}
\end{figure}
We applied our algorithm to the MNIST dataset. For the generator, which is the only neural network required by our approach, we used two fully connected layers (as mentioned full connectivity is the only structure considered in this work) with dimensions $10\times128\times784$. Parameter $h$ was selected $h=36$ and in order to exploit the parallel processing capabilities of our computational platform we used a batch size of 32. Finally we applied a smoothing factor $\lambda=0.999$ for the power estimation of each component of the gradient matrices, while for the learning rate we selected $\mu=10^{-3}$. 

\begin{figure}[h!]
\centerline{\hfill\includegraphics[width=0.495\textwidth]{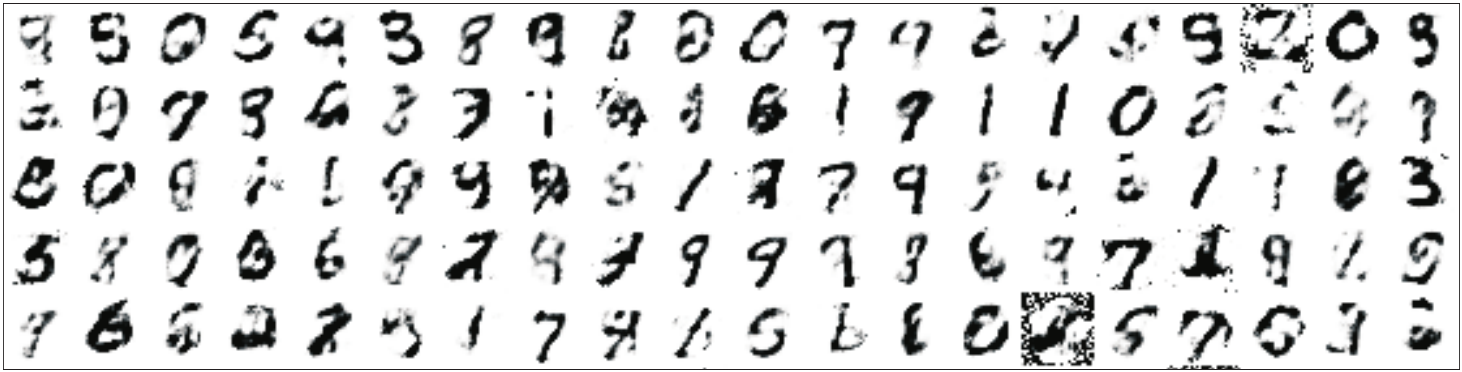}\hfill\includegraphics[width=0.495\textwidth]{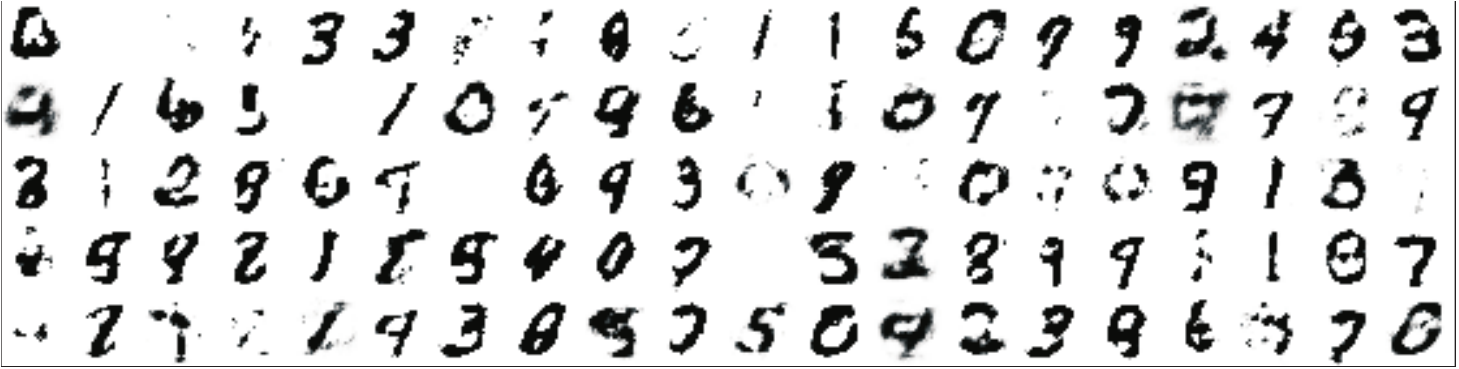}\hfill}
\vskip-0.1cm
\centerline{\hfill\hbox to 0.495\textwidth{\hfill\small(a)\hfill}\hfill\hbox to 0.495\textwidth{\hfill\small(b)\hfill}\hfill}
\caption{Training with MNIST dataset: (a)~Proposed method. (b)~GAN implementation following \cite{Arjovsky}.}
\label{fig:fig4}
\end{figure}

Similar geometry for the generator was adopted for the GAN implementation, namely two layers with dimensions $10\times128\times784$ while the discriminator structure was chosen to be $784\times128\times1$. We also used a batch size of 32 with the same $\lambda$ and $\mu$. We should mention that in this case we also employed smoothing of the average gradient as suggested in \cite{Adam} with a corresponding smoothing factor equal to $0.9$.
We plot the relative performance of the two methods in Figure\,\ref{fig:fig5} where we depict their Inception Score \cite{DBLP:journals/corr/abs-1802-03446,huang2018an} as a function of processing time. Our method attains better score values which is translated into more visually meaningful synthetic images as we can see from Figure\,\ref{fig:fig4}.

\begin{figure}[!h]
\centerline{\hfill\includegraphics[width=0.495\textwidth]{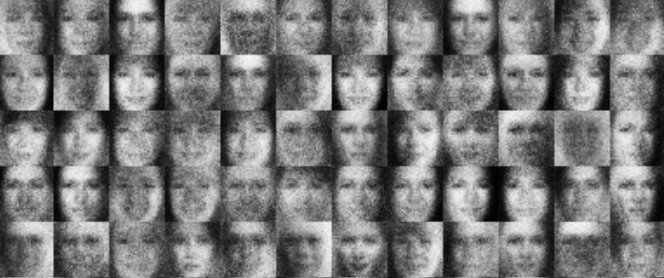}\hfill\includegraphics[width=0.495\textwidth,height=3.42cm]{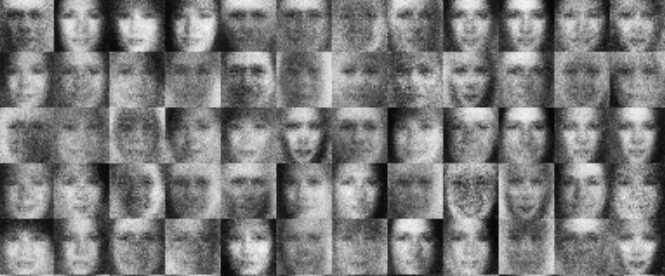}\hfill}
\vskip-0.1cm
\centerline{\hfill\hbox to 0.495\textwidth{\hfill\small(a)\hfill}\hfill\hbox to 0.495\textwidth{\hfill\small(b)\hfill}\hfill}
\vskip0.3cm
\centerline{\hfill\includegraphics[width=0.495\textwidth]{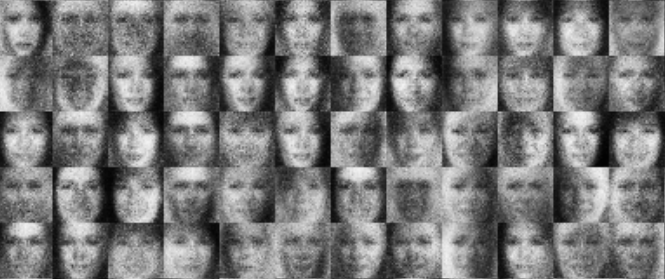}\hfill\includegraphics[width=0.495\textwidth]{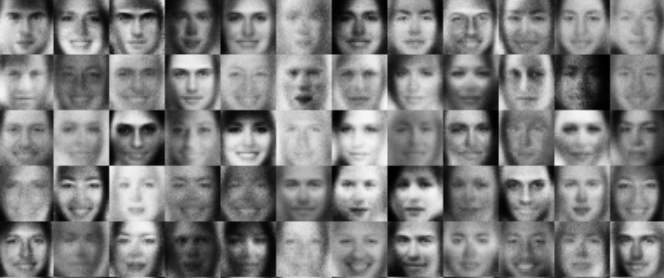}\hfill}
\vskip-0.1cm 
\centerline{\hfill\hbox to 0.495\textwidth{\hfill\small(c)\hfill}\hfill\hbox to 0.495\textwidth{\hfill\small(d)\hfill}\hfill}
\caption{Training with (cropped) CelebA dataset: GANs after (a) $10^6$ iterations; (b) $3\times10^6$ iterations; (c) $5\times10^6$ iterations; (d)~Proposed method after $5\times10^6$ iterations. Batch size $K=64$.}\label{fig:fig6}
\end{figure}

We also applied our algorithm to the CelebA dataset (properly cropped). The geometry used for the generator was again two layers with dimensions $20\times300\times1024$, $h=80$, batch size 64, $\lambda=0.999, \mu=10^{-4}$. In Figure\,\ref{fig:fig6} we present results generated by the GAN architecture after (a)~$10^6$; (b) $3\times10^6$; (c)~$5\times10^6$ iterations and batch size $K=64$, while in (d)~the results of our method after $5\times10^6$ iterations and the same batch size. We must clarify that the $5\times10^6$ iterations 
of our method that led to Figure\,\ref{fig:fig6}(d) required less physical time than the $10^6$ iterations with GANs which produced Figure\,\ref{fig:fig6}(a). As we can see, our algorithm provides far superior synthetic images even when compared to the GAN results captured in Figure\,\ref{fig:fig6}(c) and produced after the same number of iterations. The latter required five times more physical time than our method. 
\vskip0.1cm
\begin{remark}\label{rem:5}
We would like to mention that we also implemented our algorithm in \textit{Matlab}. When executed on a \textit{laptop} computer (MacBook Pro) in the case of the MNIST database, a single sweep of its 60,000 images required approximately 3\,min. When, however, we exploited the vectorized computation capacity of Matlab and developed a batched version of the algorithm, the execution time was reduced to 4\,sec with a batch of size~32.
\end{remark}
\vskip0.1cm

\begin{figure}[!h]
\vskip-0.2cm
\centerline{\hfill\includegraphics[width=0.495\textwidth,height=3.435cm]{fig6d.pdf}\hfill\includegraphics[width=0.495\textwidth]{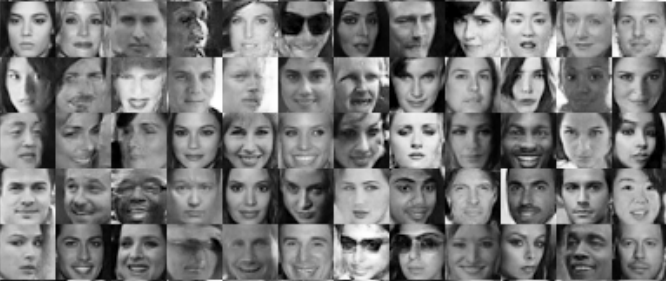}\hfill}
\vskip-0.1cm
\centerline{\hfill\hbox to 0.495\textwidth{\hfill\small(a)\hfill}\hfill\hbox to 0.495\textwidth{\hfill\small(b)\hfill}\hfill}
\caption{Training with (cropped) CelebA dataset: (a)~Proposed method after $5\times10^6$ iterations (Figure\,\ref{fig:fig6}(d) repeated); (b)~Deep convolutional GANs \cite{Radford2015UnsupervisedRL} after $200,\!000$ iterations. Batch size $K=64$.}\label{fig:fig7}
\vskip-0.3cm
\end{figure}

We can experience truly exceptional improvement in image quality with GANs if we utilize deep convolutional networks \cite{Radford2015UnsupervisedRL} instead of the fully connected version previously discussed. The quality improvement can indeed be verified from Figure\,\ref{fig:fig7} 
where in (a) we repeat the results of our method from Figure\,\ref{fig:fig6}(d) while in (b) we present the results of a deep convolutional GAN with geometry as the one suggested in \cite{Gulrajani}. We should however note that this amelioration in GANs comes at an extremely elevated computational cost. Indeed the physical time required for 200,000 iterations (with batch size $K=64$) was five times more than the 5 million iterations of our method.

It is also possible to measure quality by adopting the Maximum Mean Discrepancy score [12]. According to this index, 
better quality images produce smaller values. The scores for the three methods are: $12\times10^{-4}$ for the convolutional GANs, $296\times10^{-4}$ for the classical GANs, and finally $17\times10^{-4}$ for our approach. As we can see, these scores are in complete accordance with the visual perception of the corresponding synthetic images.

One might presume that, in our method, if we continue iterating this could lead to further improvement, possibly matching the quality of convolutional GANs. Unfortunately this is not the case due to saturation phenomena that occur when the parameter estimation method reaches its steady-state phase. Indeed the residual perturbations (explained in Section\,\ref{ssec:3.1}) prohibit any additional enhancement.

Finally, we would like to add that, currently our group is targeting the extension of our method to convolutional networks. The goal is to be able to avoid adversarial approaches by developing a non-adversarial training method which enjoys significant reduction in computational complexity and robustness in convergence. Basically, we would like the convolutional networks to inherit the same positive characteristics established for their fully connected counterparts.

\section{Acknowledgment}
This work was supported by the US National Science Foundation under Grant CIF\,1513373, through Rutgers University.

{\small
\bibliographystyle{ieee}
\bibliography{egbib}

\begin{thebibliography}{10}\itemsep=-1pt

\bibitem{Andrews}
D.~F. Andrews, R.~Gnanadesikan, and J.~L. Warner.
\newblock Transformations of multivariate data.
\newblock {\em Biometrics}, 27(4):825--840, 1971.

\bibitem{Arjovsky}
M.~Arjovsky, S.~Chintala, and L.~Bottou.
\newblock Wasserstein generative adversarial networks.
\newblock In {\em {ICML}}, volume~70 of {\em Proceedings of Machine Learning
  Research}, pages 214--223. {PMLR}, 2017.

\bibitem{Bengio-tricks-arxiv2012}
Y.~Bengio.
\newblock Practical recommendations for gradient-based training of deep
  architectures.
\newblock Technical Report Arxiv report 1206.5533, Universit{\'{e}} de
  Montr{\'{e}}al, 2012.

\bibitem{Benveniste}
A.~Benveniste and M.~Metivier.
\newblock {\em Adaptive Algorithms and Stochastic Approximations}.
\newblock Applications of Mathematics. Springer, 1990.

\bibitem{DBLP:journals/corr/abs-1802-03446}
A.~Borji.
\newblock Pros and cons of {GAN} evaluation measures.
\newblock {\em CoRR}, abs/1802.03446, 2018.

\bibitem{Box}
G.~E.~P. Box and D.~R. Cox.
\newblock An analysis of transformations.
\newblock {\em Journal of the Royal Statistical Society. Series B},
  26(2):211--252, 1964.

\bibitem{Creswell}
A.~Creswell, T.~White, V.~Dumoulin, K.~Arulkumaran, B.~Sengupta, and A.~A.
  Bharath.
\newblock Generative adversarial networks: An overview.
\newblock {\em IEEE Signal Processing Magazine}, 35(1):53--65, January 2018.

\bibitem{Cybenko}
G.~Cybenko.
\newblock Approximation by superpositions of a sigmoidal function.
\newblock {\em Mathematics of Control, Signals and Systems}, 2(4):303--314,
  1989.

\bibitem{Dziugaite}
G.~K. Dziugaite, D.~M. Roy, and Z.~Ghahramani.
\newblock Training generative neural networks via maximum mean discrepancy
  optimization.
\newblock {\em Arxiv 1505.03906}, 2015.

\bibitem{Glorot}
X.~Glorot and Y.~Bengio.
\newblock Understanding the difficulty of training deep feedforward neural
  networks.
\newblock In {\em Proceedings International Conference on Artificial
  Intelligence and Statistics}, 2010.

\bibitem{Goodfellow}
I.~J. Goodfellow, J.~Pouget-Abadie, M.~Mirza, B.~Xu, D.~Warde-Farley, S.~Ozair,
  A.~Courville, and Y.~Bengio.
\newblock Generative adversarial nets.
\newblock {\em Arxiv 1406.2661}, 2014.

\bibitem{Gretton:2012:KTT:2503308.2188410}
A.~Gretton, K.~M. Borgwardt, M.~J. Rasch, B.~Sch\"{o}lkopf, and A.~Smola.
\newblock A kernel two-sample test.
\newblock {\em J. Mach. Learn. Res.}, 13(1):723--773, Mar. 2012.

\bibitem{Gulrajani}
I.~Gulrajani, F.~Ahmed, M.~Arjovsky, V.~Dumoulin, and A.~Courville.
\newblock Improved training of \uppercase{W}asserstein \uppercase{GAN}s.
\newblock {\em arXiv:1704.00028}, 2017.

\bibitem{huang2018an}
G.~Huang, Y.~Yuan, Q.~Xu, C.~Guo, Y.~Sun, F.~Wu, and K.~Weinberger.
\newblock An empirical study on evaluation metrics of generative adversarial
  networks, 2018.

\bibitem{Adam}
D.~P. Kingma and J.~L. Ba.
\newblock \uppercase{ADAM}: A method for stochastic optimization.
\newblock {\em International Conference on Learning Representations}, 2015.

\bibitem{Li}
Y.~Li, K.~Swersky, and R.~Zemel.
\newblock Generative moment matching networks.
\newblock {\em Arxiv 1502.02761}, 2015.

\bibitem{Masters2018RevisitingSB}
D.~Masters and C.~Luschi.
\newblock Revisiting small batch training for deep neural networks.
\newblock {\em CoRR}, abs/1804.07612, 2018.

\bibitem{Mescheder}
L.~M. Mescheder, S.~Nowozin, and A.~Geiger.
\newblock The numerics of \uppercase{GAN}s.
\newblock In {\em Proceedings Advances Neural Information Processing Systems
  Conference}, 2017.

\bibitem{Radford2015UnsupervisedRL}
A.~Radford, L.~Metz, and S.~Chintala.
\newblock Unsupervised representation learning with deep convolutional
  generative adversarial networks.
\newblock {\em CoRR, abs/1511.06434}, 2015.

\bibitem{Hinton}
T.~Tieleman and G.~Hinton.
\newblock Lecture 6.5 - rmsprop.
\newblock COURSERA: Neural Networks for Machine Learning, 2012.

\end{thebibliography}
}
\appendix
\section{Appendix}
In the Appendix we provide the Matlab code for the algorithm in Table\,\ref{tab:2}. We recall that this algorithm performs training of a fully connected two-layer neural network. If one chooses to use this program, the training dataset must be in a .MAT file in the form of a single matrix called \textit{data}. The columns of this matrix must contain the training vectors. With this program we exploit Matlab's vectorized computations and propose a batched version of the algorithm appearing in Table\,\ref{tab:2}. The corresponding speedup in execution, as was mentioned in Remark\,\ref{rem:5}, is significant even for a single processor platform. We believe that with such execution times it is no longer unrealistic to perform training of generative networks on laptops using Matlab.

\noindent{\scriptsize
\begin{verbatim}
function [A,a,B,b]=kerntrain(data_file,n,m,h,mu,rounds,batch);

%% Input Parameters
% file_name: is a string containing the data file name
% n: input size
% m: first layer output size.
% h: is the parameter of the Gaussian Kernel. (Typical value 80)
% mu: learning rate. (Typical value 0.0001)
% rounds: is how many times we would like to recircle the elements of the
%         database. (Typical value 50)
% batch: batch size. (Typical value 32 or 64)
%
%% Output Parameters
% A,a: first layer parameters
% B,b: second layer parameters

%% Database loading
% Data must be in the file: data_file.mat. The file must contain a
% single matrix. This matrix must be called:     data
%
% Each matrix column corresponds to a different image reshaped into a
% (column) vector.
load(data_file)

%% RMSprop (smoothing) parameter
lambda = 0.999;
e = 10^-8; % small number to be used to avoid divisions by 0.

%% Initialization following Clorot and Bengio, 2010
times = size(data,2); % length of database
k = size(data,1); % final output size
A = randn(m,n)/sqrt(m/2); 
a = zeros(m,1);
B = randn(k,m)/sqrt((k+m)/4); 
b = zeros(k,1);
Z0 = zeros(n,batch);
S0 = zeros(m,batch);
Y0 = zeros(k,batch);
V0 = zeros(k,batch);
U0 = zeros(m,batch);
M  = zeros(k,m+1);
N  = zeros(m,n+1);

%% Main Algorithm following Table 2, (batched version)
for r=1:rounds
    ZZ = randn(n,times); % generate all inputs for one epoch
    for t = batch:batch:times
        Z = ZZ(:,t-batch+1:t); % select batch inputs
        X = data(:,t-batch+1:t); % select batch training vectors
        % Compute batch outputs of the two layers
        W = A*Z + repmat(a,1,batch);
        S = max(W,0);   % ReLU
        T = B*S + repmat(b,1,batch);
        Y = 1./(1 + exp(-T));  % Sigmoid
        % Compute gradients
        YX = Y - X;
        YY0  = Y - Y0;
        KernYX  = exp(-sum(YX.^2,1)/h);
        KernYY0 = exp(-sum(YY0.^2,1)/h);
        R = repmat(KernYX,k,1).*YX - repmat(KernYY0,k,1).*YY0;
        V = (Y - Y.^2).*R;
        U = max(sign(W),0).*(B'*V);
        G = V*[S' ones(batch,1)] + V0*[S0' ones(batch,1)];
        D = U*[Z' ones(batch,1)] + U0*[Z0' ones(batch,1)];
        % Compute average power of gradient elements
        if (t==batch)&&(r==1) % if the first time, don't smooth
            M = G.^2;
            N = D.^2;
        else  % otherwise smooth
            M = lambda*M + (1-lambda)*G.^2;
            N = lambda*N + (1-lambda)*D.^2;
        end
        % Update network parameters
        B = B - mu*G(:,1:end-1)./sqrt(M(:,1:end-1)+e);
        b = b - mu*G(:,end)./sqrt(M(:,end)+e);
        A = A - mu*D(:,1:end-1)./sqrt(N(:,1:end-1)+e);
        a = a - mu*D(:,end)./sqrt(N(:,end)+e);
        % Update variables needed for the next iteration
        Z0 = Z;
        S0 = S;
        Y0 = Y;
        V0 = V;
        U0 = U;
    end
end
\end{verbatim}
}

\end{document}